\documentclass{article}

\usepackage[final]{neurips_2024}
\usepackage[utf8]{inputenc} 
\usepackage[T1]{fontenc}    
\usepackage{hyperref}       
\usepackage{url}            
\usepackage{booktabs}       
\usepackage{amsfonts}       
\usepackage{nicefrac}       
\usepackage{microtype}      
\usepackage{xcolor}         
\usepackage{graphicx}
\usepackage{amsmath}
\usepackage{subcaption}

\makeatletter
\renewcommand{\@noticestring}{}
\makeatother

\title{Reproducing AlphaZero on Tablut: Self-Play RL for an Asymmetric Board Game}

\author{%
  Tõnis Lees \\
  Institute of Computer Science\\
  University of Tartu\\
  \texttt{tonis.lees@gmail.com}
  \And
  Tambet Matiisen\thanks{Supervisor} \\
  Institute of Computer Science\\
  University of Tartu\\
  \texttt{tambet.matiisen@ut.ee}
}

\begin{document}

\maketitle

\section{Introduction}
\citet{silver2018general} introduced $AlphaZero$, a general reinforcement learning algorithm that masters Chess, Shogi, and Go through self-play with no domain-specific knowledge beyond the rules. A single neural network $(\mathbf{p}, v) = f_\theta(s)$ is trained iteratively: each iteration generates games via MCTS-guided self-play, then updates $\theta$ to better predict game outcomes and the search-refined policy. In the original formulation, both players share a single policy and value head, and positions are canonicalized so the current player's pieces always appear as ``friendly'', which is natural for symmetric games where both sides face structurally identical decisions. In asymmetric games, where players differ in piece counts, objectives, and win conditions, this single head must learn two distinct evaluation functions, which can hinder learning efficiency and performance.

This work investigates whether $AlphaZero$'s self-play framework transfers to such a setting by applying it to Tablut, a historical board game played on a $9\times9$ board with 16 attackers against 8 defenders and a king. The attacker aims to capture the king; the defender aims to escort it to a corner (see Appendix~\ref{app:rules}). The core algorithm transfers with one key modification—separate policy and value heads per player—but the asymmetric structure introduces training instabilities, such as catastrophic forgetting between roles, that were not reported in the original work.

\section{Reproduction}

The neural network architecture closely follows \cite{silver2018general}, with the main difference being a reduced residual trunk of 8 blocks with 128 filters to match Tablut’s lower complexity and more limited compute budget. The key modification is the use of separate policy and value heads for each player, $(\mathbf{p_{a}}, v_a, \mathbf{p_{d}}, v_d) = f_\theta(s)$, motivated by the fundamentally different objectives — king capture versus king escape — and the resulting asymmetric evaluation landscape. During MCTS, the head corresponding to the current player is selected, while the shared residual trunk learns common board features such as piece mobility and capture threats. Tablut's rook-like piece movement allows direct reuse of the action encoding from \cite{silver2018general}: each of the 81 squares has 32 directional planes (8 distances $\times$ 4 directions), yielding an action space of 2592 possible moves per position. See Appendix~\ref{app:state_repr} for state representation planes.

The system was implemented in JAX using Flax, Optax, Mctx, and Flashbax \citep{jax2018github, deepmind2020jax, flashbax}. To enable hardware-accelerated self-play, game environments were fully vectorized. While \citet{koyamada2023pgx} developed the Pgx library for this purpose, it does not natively support Tablut. Consequently, Tablut game logic was implemented from scratch by extending the Pgx base framework. Training was conducted on 2 NVIDIA H200 GPUs for 100 self-play iterations. In each iteration, generated states were stored in a replay buffer along with the MCTS-refined policy and the outcome reward of the corresponding game. The network was then trained on batches sampled from this buffer to predict game outcomes and policy distributions. MCTS was performed using the $Gumbel MuZero$ variant \citep{danihelka2022policy} with 128 simulations per move, which provides a more sample-efficient search than regular $PUCT$ under limited compute budget, while retaining the $AlphaZero$-style policy and value targets. The loss function combines mean squared error for value and cross-entropy for policy, as in $AlphaZero$. $AdamW$ with weight decay regularization was used for optimization. See Appendix~\ref{app:hyperparameters} for all hyper-parameters.

During training, performance against earlier checkpoints degraded — a phenomenon known as catastrophic forgetting in self-play. To stabilize training, C$_4$ data augmentation was applied (random $90^\circ$ board rotations), and the replay buffer was increased from 8 to 16 self-play iterations. Inspired by \cite{vinyals2019grandmaster}, 25\% of training games were played against randomly sampled past checkpoints, with the current model alternating between attacker and defender roles. See Appendix~\ref{app:elo_comparison} for an ablation of these contributions.

Starting from the 20th iteration, the model was evaluated every 5 iterations against 4 opponents randomly sampled from a pool of up to 10 past checkpoints. The randomly initialized model (iteration 0) was retained as a fixed anchor. Following \cite{silver2018general}, the $BayesElo$ program was used to calculate the Elo ratings across iterations. $BayesElo$ models draw rates and first-mover advantage, which is important for an asymmetric game where the inherent balance between sides is unknown.

\section{Results and discussion}

\begin{figure}[ht!]
     \centering
     \begin{subfigure}[b]{0.48\textwidth}
         \centering
         \includegraphics[width=\textwidth]{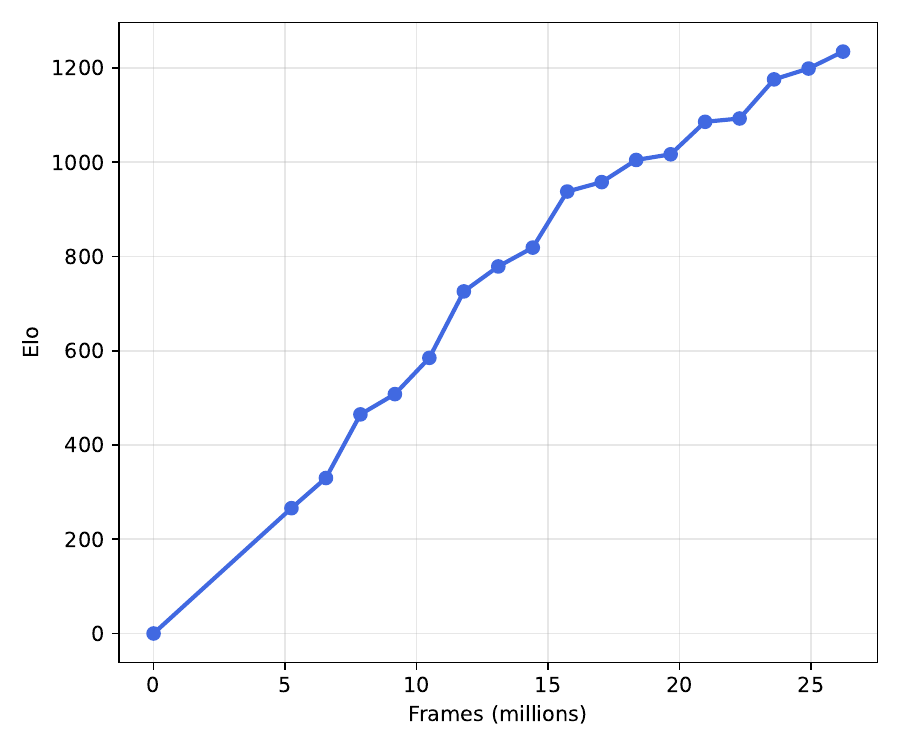}
         \caption{Elo rating progression.}
         \label{fig:elo}
     \end{subfigure}
     \hfill
     \begin{subfigure}[b]{0.48\textwidth}
         \centering
         \includegraphics[width=\textwidth]{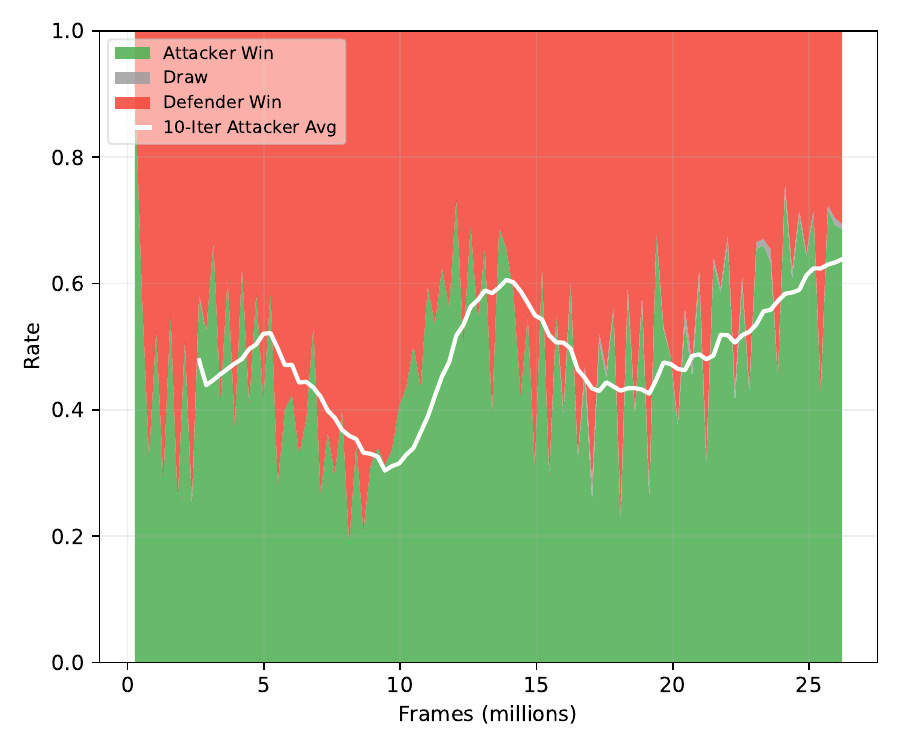}
         \caption{Game outcomes (win rates).}
         \label{fig:outcomes}
     \end{subfigure}
     
     \caption{Training metrics over 100 iterations. The model shows steady Elo growth and an increasing win rate for the attacker in Tablut.}
     \label{fig:training_results}
\end{figure}

Over 100 iterations (~26 million frames), the model reached a BayesElo rating of 1235 relative to the randomly initialized baseline (see Figure~\ref{fig:elo}). This indicates steady improvement, though the rating is only meaningful relative to this internal pool and not comparable to $AlphaZero$'s absolute performance on Chess or Go. Policy entropy decreased from 3.05 to 1.47 and the average number of pieces remaining at game end dropped from 22 to 15 (see Appendix~\ref{app:stats}), reflecting increasingly focused and decisive play. Up to iteration 75, attackers and defenders achieved similar evaluation win rates of roughly 70--80\% against the pool of past checkpoints when playing as that side. After that point, the defender’s win rate declined to 52\%, while the attacker’s climbed to 86\%, and in self-play the attacker’s 10-iteration rolling average win rate reached about 65\% by iteration 100 (see Figure~\ref{fig:outcomes}). This suggests either that the Tablut ruleset used favours the attacker or that defender strategies are harder to learn in this setting; a separate balance analysis was not performed to distinguish these explanations.

The most challenging aspects of the reproduction were (i) down-scaling training from the much larger compute used by \citet{silver2018general} and (ii) mitigating catastrophic forgetting in self-play, which the original paper does not discuss in detail. Among the hyperparameters, the MCTS simulation count proved most critical: reducing it below 128 degraded both search quality and the resulting policy targets, while larger counts were impractical on two GPUs. Increasing the replay buffer size and using C$_4$ data augmentation both substantially improved stability.

\bibliographystyle{plainnat}
\bibliography{references}

\newpage
\appendix

\section{Appendix}

\subsection{Tablut rules}
\label{app:rules}
Tablut is a historic board game that was played in Northern Europe and its rules were first documented by Carl Linnaeus in 1732 in his diary. The original text, written in a mixture of Latin and Swedish, was subject to several problematic translations, leading to many different rule interpretations today. The rules used in this replication are the following:
\begin{itemize}
    \item The game is played on a 9x9 board with 8 defenders, 1 king and 16 attackers. Regular pieces are also called taflmen.
    \item The king starts on the center square which is called the throne (see Figure~\ref{fig:board}). 
    \item The pieces move like rooks in chess, and they cannot pass through other pieces.
    \item The objective of the defenders is to move the king to any of the corners of the board.
    \item The objective of the attackers is to capture the king.
    \item Any piece can be captured by sandwiching it between two enemy pieces. If a piece moves between two enemy pieces itself, then it is not captured.
    \item The king is captured like any other piece and it can also participate in captures.
    \item Only the king can step on the corners and the throne, but other pieces may pass through the throne.
    \item The throne and the corners are hostile squares, meaning that they can be used to capture pieces by both players. The throne becomes hostile to the defenders only when the king is not in the throne.
    \item If no captures have been made in the past 100 moves or it has not ended in 512 moves, then the game is a draw.
    \item If any board state appears for the third time during the game then the player who made the third repetition loses.
    \item If a player has no legal moves left, then they lose.
\end{itemize}
The game starts from a certain piece placement:
\begin{figure}[ht!]
    \centering
    \includegraphics[width=0.4\linewidth]{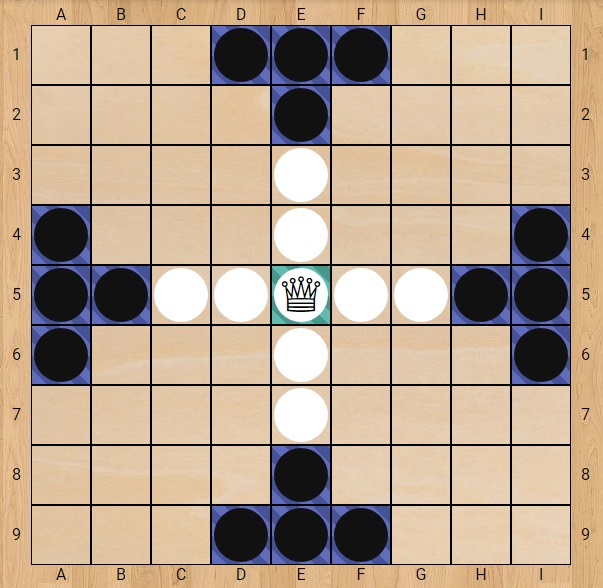}
    \caption{Tablut board's initial state}
    \label{fig:board}
\end{figure}

\newpage
\subsection{Elo Progression Ablation}
\label{app:elo_comparison}

\begin{figure}[ht!]
\centering
\includegraphics[width=0.7\textwidth]{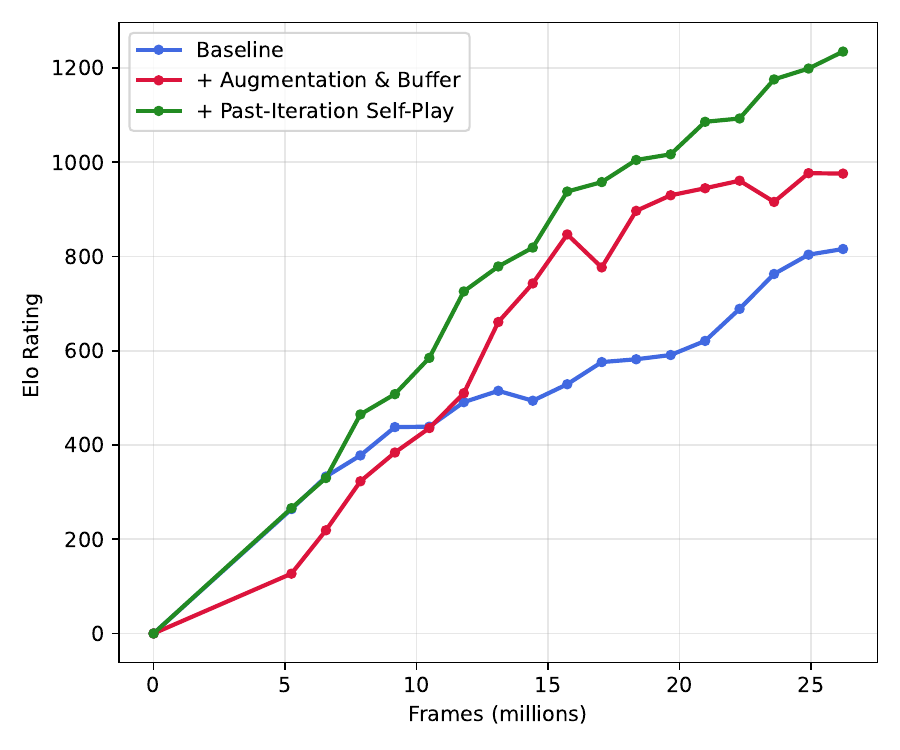}
\caption{Elo progression across three cumulative training configurations. Each successive run adds one stabilization technique to the previous configuration.}
\label{fig:elo_comparison}
\end{figure}

Figure~\ref{fig:elo_comparison} compares the Elo progression across three training configurations, each building cumulatively on the previous one. The \textit{Baseline} run uses the dual-head AlphaZero architecture with a replay buffer of 8 self-play iterations and no data augmentation. The \textit{+ Augmentation \& Buffer} run adds C$_4$ data augmentation (random $90^{\circ}$ board rotations) and increases the replay buffer from 8 to 16 self-play iterations ($16 \times 1{,}024 \times 256 \approx 4.2$M states). The \textit{+ Past-Iteration Self-Play} run further adds 25\% of training games played against randomly sampled past checkpoints. Each technique yields a clear improvement in final Elo rating, with the full configuration reaching 1235.

\subsection{State Representation and Hyperparameters}
\label{app:state_repr}

The input state $s$ is a $9 \times 9 \times 43$ tensor, summarized in
Table~\ref{tab:state_repr}.

\begin{table}[ht!]
\centering
\caption{State representation planes.}
\label{tab:state_repr}
\begin{tabular}{llr}
\toprule
\textbf{Feature} & \textbf{Description} & \textbf{Planes} \\
\midrule
\multicolumn{3}{c}{\textit{Per history step ($\times$ 8 steps)}} \\
Friendly pieces & Current player's taflmen & 1 \\
Enemy pieces & Opponent's taflmen & 1 \\
King & King position & 1 \\
Repetition $\geq 1$ & Position seen before & 1 \\
Repetition $\geq 2$ & Position seen twice & 1 \\
\midrule
\multicolumn{3}{c}{\textit{Auxiliary}} \\
Player color & Current side to move & 1 \\
Total move count & Normalized step count & 1 \\
Half-move clock & Moves since last capture & 1 \\
\midrule
\textbf{Total} & & \textbf{43} \\
\bottomrule
\end{tabular}
\end{table}

\label{app:hyperparameters}
The hyperparameters used for the Tablut $AlphaZero$ replication are summarized in Table~\ref{tab:hyperparameters}.

\begin{table}[ht!]
\centering
\caption{Hyperparameters used for training and MCTS.}
\label{tab:hyperparameters}
\begin{tabular}{ll}
\toprule
\textbf{Hyperparameter} & \textbf{Value} \\
\midrule
\multicolumn{2}{c}{\textit{Neural Network Architecture}} \\
Residual blocks & 8 \\
Filters & 128 \\
Policy heads & 2 (Attacker, Defender) \\
Value heads & 2 (Attacker, Defender) \\
\midrule
\multicolumn{2}{c}{\textit{Training \& Optimization}} \\
Optimizer & $AdamW$ \\
Weight decay & 0.0001 \\
Learning rate schedule & Cosine decay with warmup \\
Warmup steps & 500 \\
Peak learning rate & 0.002 \\
Minimum learning rate & 0.00001 \\
Total training steps & 102,400 \\
Batch size & 512 \\
\midrule
\multicolumn{2}{c}{\textit{Self-Play \& MCTS}} \\
Number of iterations & 100 \\
Parallel games per iteration & 1024 \\
Steps per game per iteration & 256 \\
MCTS simulations per move & 128 \\
Replay buffer size & 4.2M states \\
Self-play vs. past versions & 25\% \\
\bottomrule
\end{tabular}
\end{table}

\newpage
\subsection{Detailed Training Statistics}
\label{app:stats}

\begin{figure}[ht!]
    \centering
    \begin{subfigure}[b]{0.48\textwidth}
        \centering
        \includegraphics[width=\textwidth]{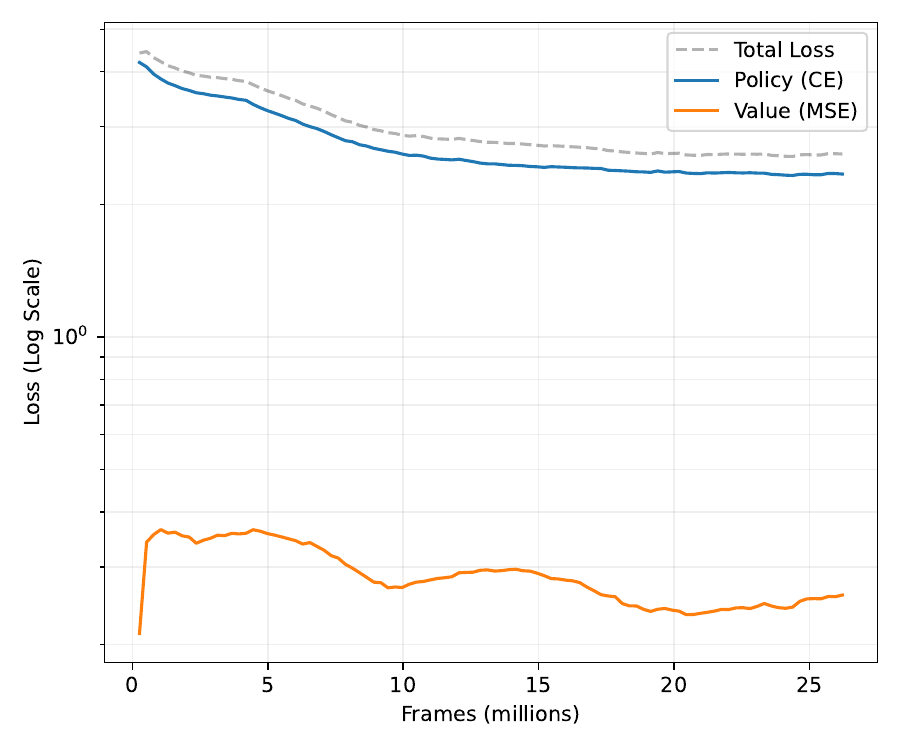}
        \caption{Combined Training Losses}
    \end{subfigure}
    \hfill
    \begin{subfigure}[b]{0.48\textwidth}
        \centering
        \includegraphics[width=\textwidth]{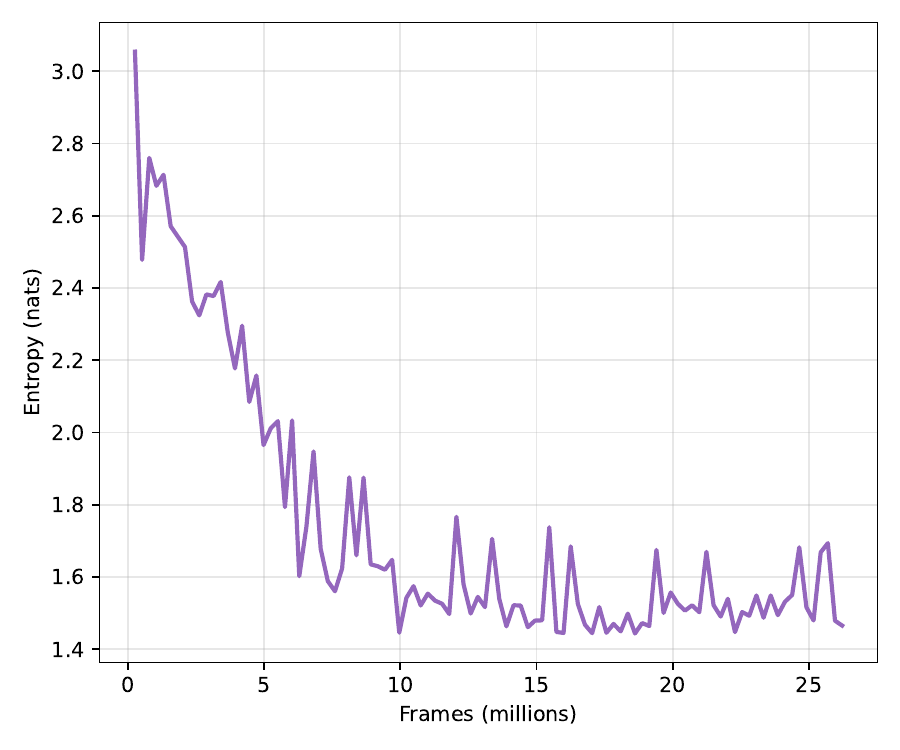}
        \caption{Policy Entropy}
    \end{subfigure}
    
    \vspace{0.3cm}
    
    \begin{subfigure}[b]{0.48\textwidth}
        \centering
        \includegraphics[width=\textwidth]{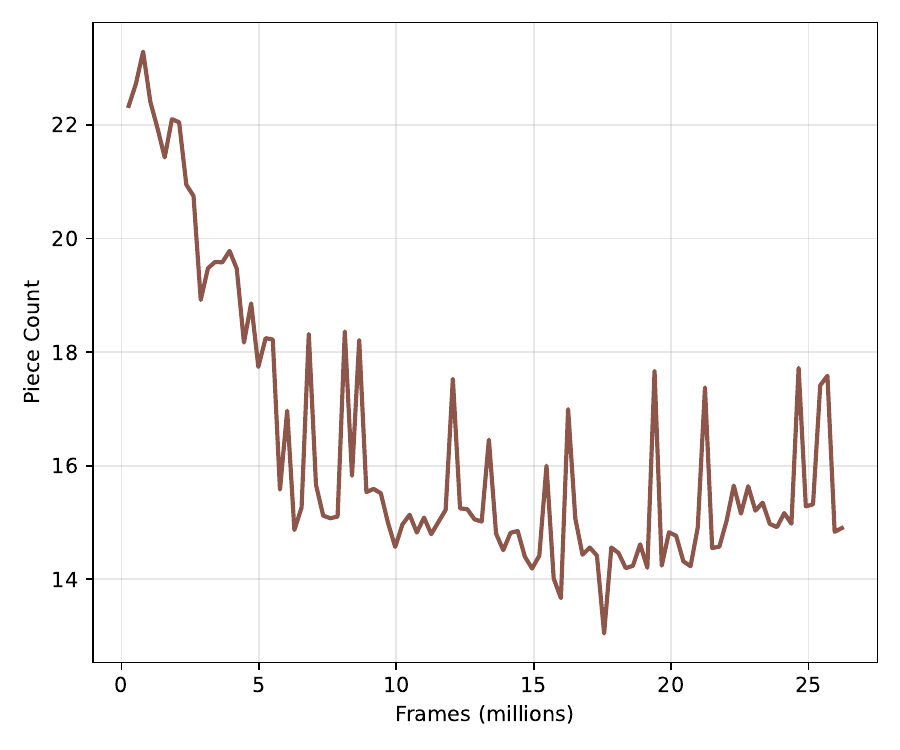}
        \caption{Average Pieces Remaining}
    \end{subfigure}
    
    \caption{Detailed training metrics and game statistics recorded over 100 self-play iterations.}
    \label{fig:detailed_metrics}
\end{figure}

Figure~\ref{fig:detailed_metrics} provides a granular view of the training process. The convergence of the policy and value losses (a) corresponds with the steady decrease in policy entropy (b), indicating the model is becoming more confident in its moves. Furthermore, the decrease in pieces remaining at the end of games (c) suggests more decisive play and efficient captures as training progresses.

\end{document}